\documentclass[sigconf]{acmart}

\usepackage{graphicx}
\usepackage{subcaption}

\AtBeginDocument{%
  }

\setcopyright{acmlicensed}
\copyrightyear{2024}
\acmYear{2024}
\acmDOI{XXXXXXX.XXXXXXX}

\acmConference[HAI '24]{Make sure to enter the correct
  conference title from your rights confirmation email}{June 03--05,
  2018}{Woodstock, NY}
\acmISBN{978-1-4503-XXXX-X/18/06}

\begin{document}

%\title{Automated Assessment and Adaptive Feedback for Psychomotor Skills Training in Quadrotor Teleoperation}
\title[Automated Assessment and Feedback Improves Quadrotor Training Outcomes]{Automated Assessment and Adaptive Multimodal Formative Feedback Improves Psychomotor Skills Training Outcomes in Quadrotor Teleoperation}

\author{Emily Jensen}
\email{emily.jensen@colorado.edu}
\orcid{0000-0002-1550-379X}
\affiliation{%
  \institution{University of Colorado Boulder}
  \city{Boulder}
  \state{Colorado}
  \country{USA}
}

\author{Sriram Sankaranarayanan}
\orcid{0000-0001-7315-4340}
\email{srirams@colorado.edu}
\affiliation{%
  \institution{University of Colorado Boulder}
  \city{Boulder}
  \state{Colorado}
  \country{USA}
}

\author{Bradley Hayes}
\email{bradley.hayes@colorado.edu}
\orcid{0000-0002-0723-1085}
\affiliation{%
  \institution{University of Colorado Boulder}
  \city{Boulder}
  \state{Colorado}
  \country{USA}
}

\renewcommand{\shortauthors}{Jensen et al.}

\begin{abstract}
  The workforce will need to continually upskill in order to meet the evolving demands of industry, especially working with robotic and autonomous systems. Current training methods are not scalable and do not adapt to the skills that learners already possess. In this work, we develop a system that automatically assesses learner skill in a quadrotor teleoperation task using temporal logic task specifications. This assessment is used to generate multimodal feedback based on the principles of effective formative feedback. Participants perceived the feedback positively. Those receiving formative feedback viewed the feedback as more actionable compared to receiving summary statistics. Participants in the multimodal feedback condition were more likely to achieve a safe landing and increased their safe landings more over the experiment compared to other feedback conditions. Finally, we identify themes to improve adaptive feedback and discuss and how training for complex psychomotor tasks can be integrated with learning theories.
\end{abstract}




\newcommand{\todo}[1]{\textcolor{red}{TODO: #1}}

\maketitle

\section{Introduction}

Future industrial development will depend on collaboration between humans and automated systems. While some fear losing jobs to automation, experts argue there will be a need for highly-skilled human-automation teams that can adapt to customer-specific tasks \cite{bell_employment_1985, moya_augmented_2023}. Humans in these collaborative teams must be able to understand how the autonomous system works, how to manage it, and how to adapt when maintenance is needed or other technical issues arise \cite{gregory_work_2000}. A recent report estimates that one third of job requirements will require technological skills that are not yet considered crucial~\cite{schwab_future_2020}, meaning that employees will need to continually adapt as technological innovation continues. 

Some of the most significant barriers to achieving this industrial development are the ability to scale up capacity as well as upskill and reskill the current workforce \cite{mukherjee_identification_2023}. Experts estimate that 50\% of existing employees will need to be retrained or upskilled by 2025 to keep up with technological advancement, placing significant pressure on both employers and employees to meet these demands~\cite{schwab_future_2020, li_reskilling_2022}. With the recent developments of artificial intelligence capabilities, researchers are considering how to improve and automate this crucial training.

Training systems and programs for developing industrial skills are a promising opportunity to expand the workforce. For example, sub-baccalaureate training programs and stackable certifications may allow disadvantaged workers to access the training needed to enter highly-skilled industrial sectors \cite{andrew_analysis_2020}. In order to achieve this goal, training programs will need to focus on transferable skills and present interfaces that are ``customizable, individualized, and on-demand'' to address the needs of each unique learner \cite{hutson_rethinking_2023}.

Current training methods such as individualized instruction and pre-recorded modules cannot scale up to meet this need to upskill. They also ignore the fact that many employees enter training with skills that can be transferred to a new task. Intelligent Tutoring Systems are designed to meet just these demands in classrooms by developing personalized models of students and building on knowledge the student has already mastered. Although previous work discusses applying these approaches outside the classroom~\cite{santos_training_2016}, existing approaches for training physical tasks has not been systematically researched and integrated with learning theory.

In this work, we develop a system that provides formative feedback on a quadrotor landing task using automated assessment from temporal logic task specifications and generative artificial intelligence. We demonstrate that foundational research in feedback can be transferred from school-based educational technology to develop adaptive training systems for complex, multi-objective interaction tasks between humans and autonomous systems.

\section{Related Work}
\label{sec:related-work}

This work focuses on training for complex psychomotor tasks. We define complex tasks as those with high variability, requiring multiple steps to complete, and whose performance depends on multiple factors \cite{williamson_automated_2006}. Psychomotor tasks require the coordination of physical (grasping, teleoperating) and cognitive (planning, decision making) elements to successfully complete the task \cite{Nicholls2014}.

\subsection{Automated Assessment for Complex Psychomotor Tasks}
 Adaptive training systems must be able to automatically assess performance before providing feedback. This is especially difficult for complex psychomotor tasks because successful performance depends on a variety of factors. 
 
 Assessment is also highly dependent on the task domain; as such, previous work in automated assessment has developed specialized methods for the specific domain. For example, Rauter et al. analyzed performance on a rowing task by comparing the velocity profile of the rowing stroke against expert performance \cite{rauter_when_2019}. Other studies similarly compare performance to expert trajectories as a benchmark for successful performance \cite{sheehan_formative_2019, davaris_importance_2019}. Surgical robotics studies have used physiological metrics such as smoothness, motion amplitudes, and muscular activation \cite{wang_toward_2018} in addition to response time for unanticipated events \cite{yang_adaptive_2024}. A recent study evaluated performance in human-robot teaming using number of collisions, number of re-grasps, and total task time \cite{perez-darpino_experimental_2023}. The metrics presented here are largely \textit{outcome-based}, meaning they provide an overall indication of task success, but lack a nuanced description of the learner's \textit{process} of completing the task.

 Additionally, the methods used to assess performance are not scalable to task variants or new domains. Recent examples of automated assessment (or more simply, error detection) include domains such as table tennis \cite{mat_sanusi_2021}, martial arts \cite{echeverria_2021, echeverria_2021_ws, portaz_exploring_2024}, piano \cite{moringen2021optimizing}, medical first responders \cite{pretolesi_ai-supported_2024}, industrial production \cite{haslgrubler_2019}, and surgery \cite{rosen_objective_2011}. Many of these methods rely on neural network classifiers, which require significant data to train and do not provide explanations for their predictions. In this work, we build off Jensen et al.'s proposed framework, which describes skill as a vector of several performance outcomes \cite{jensen_more_2023}. They assess performance relative to a set of task specifications, which are easy to compute and require no data to learn. Task specifications use logical requirements and constraints to define task objectives, such as reaching the goal state within $T$ time: $\textsc{Eventually}_{[0,T]} atGoal$.

\subsection{Formative Feedback}
\label{sec:rw-formative-feedback}
Providing feedback is key to improving a learner's performance. Summative feedback provides a general summary of performance after the learning program is completed \cite{steinfeld_common_2006}. While useful for providing an overview of performance, learners are left to self-regulate their practice in the absence of other feedback. On the other hand, formative feedback is provided during the learning process to help guide future learning \cite{shute_focus_2008}. This type of feedback is given more frequently and focuses on encouragement. Based on recent reviews in the educational technology literature, we identified the following elements of effective formative feedback:
\begin{itemize}
    \item \textit{Reflection}: feedback gives detailed information about the task, process, and encourages the learner to self-reflect \cite{pishchukhina_supporting_2021, martinez_content-focused_2023, banihashem_systematic_2022, henderson_conditions_2019}.
    \item \textit{Motivation}: feedback expresses confidence in the learner’s abilities \cite{pishchukhina_supporting_2021, martinez_content-focused_2023, hatziapostolou_enhancing_2010}.
    \item \textit{Timely}: feedback is directly connected to the learner's recent actions \cite{martinez_content-focused_2023, hatziapostolou_enhancing_2010}.
    \item \textit{Actionable}: feedback provides specific guidance for improvement that is related to the assessment criteria \cite{martinez_content-focused_2023, hatziapostolou_enhancing_2010, henderson_conditions_2019}.
    \item \textit{Manageable}: feedback is detailed but not overwhelming to interpret \cite{hatziapostolou_enhancing_2010, henderson_conditions_2019}.
\end{itemize}

These elements of feedback have been shown to support learning outcomes and are positively perceived by students in classroom learning settings. One goal of this study is to evaluate whether this theory of effective feedback improves task performance in a complex psychomotor task domain.

\subsection{Training for Psychomotor Tasks}

Recent work in developing end-to-end adaptive training systems for complex psychomotor tasks has focused on individual domains such as surgery, sports, marksmanship, karate, driving, aircraft maintenance, and additive manufacturing \cite{laine_systematic_2022, zotov_moving-target_2023, santos_training_2016}. 

Several works have discussed how pedagogically-informed feedback strategies may be implemented in training systems. For example, Korhonen et al. \cite{korhonen_training_2023} and P\'erez-Ram\'irez et al. \cite{perez-ramirez_intelligent_2016} discuss how theories such as embodied cognition can be implemented into virtual reality learning environments. Other work proposes inserting erroneous solutions to encourage critical thinking \cite{buche_mascaret_2003} or using adaptive epistemic feedback for training \cite{luengo_target_2013}. However, none of these studies have implemented and evaluated the effectiveness of these theories.

Training systems that provide performance feedback tend to rely on prerecorded responses or templates for reacting to failure modes \cite{di_mitri_keep_2022, ortiz_capturing_2020, davaris_importance_2019} or display statistical summaries of key performance outcomes \cite{zotov_moving-target_2023, sheehan_formative_2019, rauter_when_2019}. An ultrasound placement study generated a visual comparison between the learner's placement and orientation compared to an expert \cite{sheehan_formative_2019}. To the best of our knowledge, no studies have investigated the use of generated natural-language text to provide formative feedback to learners. In this study, we provide statistical summaries as a baseline condition and compare learner performance to generated text containing the identified elements of effective formative feedback.

\section{Contributions and Research Questions}
The contribution of this paper is \textbf{a flexible and validated framework for automatically assessing performance in multi-objective tasks and generating personalized formative feedback}. We accomplish this by pairing robustness measures of formal task specifications with natural language feedback generated from pedagogically-grounded templates. In this study, we compare groups that received summary metrics of their performance, automatically generated text feedback, and text feedback paired with an annotated figure showing their trajectory. We evaluate the system using the following research questions (RQs).
\begin{itemize}
    \item \textit{RQ 1}. Do participants perceive the elements of formative feedback differently between score-based, semantic, or multimodal presentations?
    \item \textit{RQ 2}. What factors predict the perception of formative feedback?
    \item \textit{RQ 3}. How does automated formative feedback affect participants' learning trajectory?
\end{itemize}

\section{Methods}

The study code and analysis scripts will be added as a link here in the final version of the paper.

\subsection{Quadrotor Landing Task}
\begin{figure}
    \centering
    \includegraphics[width=1\linewidth]{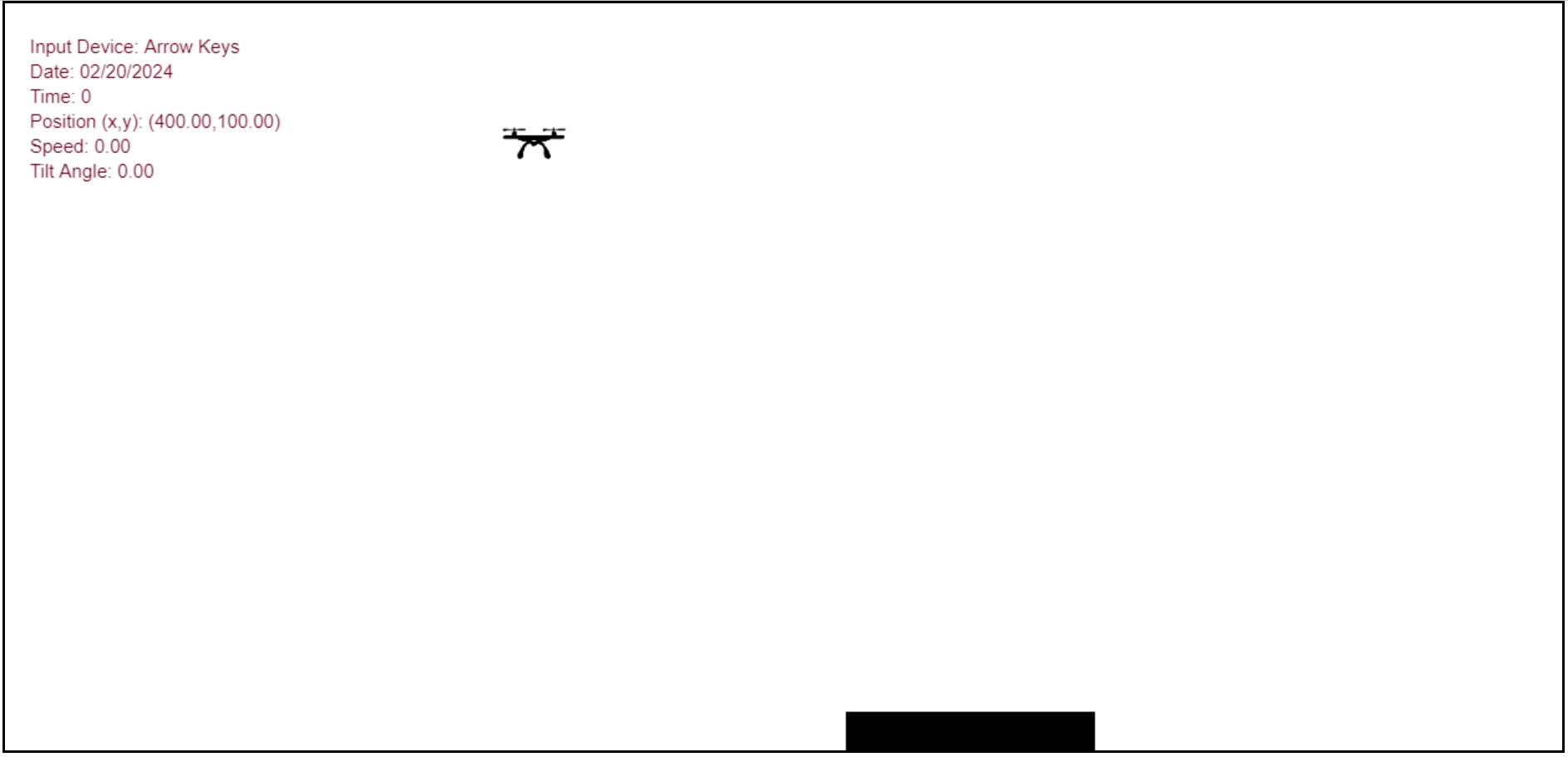}
    \Description{A screenshot of a rectangular simulation area. Red text at the top left corner gives information about velocity and angle. A black 2D drone is positioned above and to the left of a black landing pad.}
    \caption{Screenshot of the starting configuration of the quadrotor landing task.}
    \label{fig:quadrotor-game}
\end{figure}

In this experiment, participants completed a simulated quadrotor landing task. In this task, participants used keyboard inputs to adjust the quadrotor's throttle (vertical force) and attitude (rotation for horizontal force). To achieve a safe landing, the quadrotor must reach the landing pad with a speed less than 15 $m/s$ and a rotation angle within $\pm 5^{\circ}$. We labeled a landing attempt as unsafe if the drone reached the landing pad but did not satisfy the required speed or angle constraints. All other landing attempts were crashes. We refer to a landing attempt as a \textit{trial}. Each trial was capped at 120 seconds. Figure~\ref{fig:quadrotor-game} shows the initial configuration of the task participants completed. The initial position of the drone and the landing pad did not change between trials. Yuh et al.'s work provides more details on the dynamics of the quadrotor and design of the simulator \cite{yuh_using_2024}.

\subsection{Participants}
We recruited participants using the Prolific platform. All participants were United States residents. 177 participants completed the study. Of these participants, 16 restarted the study due to technical issues. To minimize confounding learning effects, we excluded six participants that completed more than five trials before restarting. We excluded another four participants who did not provide a good faith effort in the experiment, as measured by never using the horizontal input controls and crashing the quadrotor on each trial. This resulted in a final dataset of 167 participants.

Participants ranged in age from 18 to 74 years, with a median age of 35 years. Their reported gender identities were 73 Men (44\%), 87 Women (52\%) and 7 Non-binary individuals (4\%). 97\% of participants reported no prior experience flying drones or have flown a drone a few times. Participants reported a range of video game experience, with 30 not playing video games (18\%), 46 playing monthly (27.5\%), 40 playing weekly (24\%) and 51 playing daily (30.5\%). 

\subsection{Experiment Design and Procedure}
We conducted a between-subjects study with three experimental conditions. In the baseline condition, participants received summary statistics such as their landing outcome and an overall score of their performance, replicating prior work \cite{yuh_classification_2024}. In the second condition, participants received AI-generated text feedback, described in Section~\ref{sec:feedback-design}. In the third condition, participants received AI-generated text feedback along with an annotated image of their trajectory, which highlighted an area of their trajectory to focus on improving. In our final dataset, 55 participants were in the baseline condition, 56 participants were in the text feedback condition, and 56 participants were in the multimodal feedback condition.

After consenting to participate in the study and reading the instructions for the task, participants completed the quadrotor landing task. After each trial, participants received feedback on their performance depending on their experimental condition. Participants then rated the feedback they received and completed the landing task again. After completing the task 20 times, participants completed a brief demographic questionnaire and rated their overall perception of the feedback they received. On average, participants completed the experiment in 29.12 minutes (SD = 10.24 minutes). They spent an average of 28.12 seconds (SD = 11.08 seconds) reviewing and rating their feedback on each trial.

\subsection{Automated Assessment}
\label{sec:automated-assessment}
The system automatically assessed landing performance using a previously validated framework \cite{jensen_more_2023}. For each component of the task, we defined a specification using signal temporal logic~\cite{Donze+Maler/2010/Robust}, a formalism for specifying complex temporal tasks. For the quadrotor landing task, the specifications focused on the safety and landing behaviors. The specifications are shown in Table~\ref{tab:specifications}. Robustness values are a quantitative score that describes how well the trajectory of the quadrotor meets the given specification; large positive values indicate better compliance (e.g., staying far away from the edge of the simulation window) while large negative values indicate stronger violations (e.g., extreme landing angle) \cite{Donze+Maler/2010/Robust,jensen_more_2023}. 
%% SS: OK to delete this
%%Due to the nature of the simulation, the safety components have a minimum robustness of 0 since the task ends if the quadrotor crashes into the side of the simulation window.

\begin{table}[ht]
\caption{Overview of specifications for quadrotor landing task with range of possible robustness values for the individual components. Note that the specific values for $s_i$ and $l_i$ depend on the size of the simulation window and the quadrotor.}
\label{tab:specifications}
\centering
\begin{tabular}{lll}
\toprule
Description & Specification & Robustness Range \\
\midrule
Avoid left edge & $s_1 = x > 0$ & [0, 1210] \\
Avoid right edge  & $s_2 = x < 1250$ & [0, 1210] \\
Avoid bottom edge & $s_3 = y > 0$  & [0, 575] \\
Avoid top edge    & $s_4 = y < 600$  & [0, 575] \\
\midrule
Avoid left land edge  & $l_1 = x > 650$  & [-650, 560] \\
Avoid right land edge & $l_2 = x < 850$  & [-360, 850] \\
Slow landing speed    & $l_3 = ||v|| < 15$ & [-17, 15] \\
Shallow landing angle  & $l_4 = |\phi| < 5$ & [-24, 5] \\
\midrule
Safety component  & $S = \wedge_{i=1}^4 s_i$ &                  \\
Landing component & $L = \wedge_{i=1}^4 l_i$ &                  \\
Complete task in 120s & $S~\textsc{until}_{[0,120]}~L$ &     \\
\bottomrule
\end{tabular}

\end{table}

To keep the feedback manageable, we used a heuristic for selecting the top area of improvement the participant should focus on for the next trial. The safety components were given the highest priority; if the quadrotor crashed into any of the sides of the simulation window (indicated by $s_i = 0$), this was selected as the area of improvement. If the quadrotor landed unsafely, either landing speed or angle was chosen as the area of improvement ($l_3$ or $l_4 < 0$). For successful landings, the area of improvement was selected as overall efficiency if the trial was longer than a predetermined length or otherwise defaulted to smoothness.

\subsection{Formative Feedback Design}
\label{sec:feedback-design}
Participants received formative feedback based on the context generated from the automated assessment in Section~\ref{sec:automated-assessment} and natural language generated from a prompt incorporating the elements of formative feedback discussed in Section~\ref{sec:rw-formative-feedback}. The prompt included a description of the target task, the identified area of improvement, the generated image of the trajectory, and an explanation of what each element of the feedback should contain. We used GPT-4V \cite{openai_gpt-4vision_2023} to generate the text feedback.

The visual feedback consisted of an image of the landing trajectory with a superimposed circle to highlight a specific area of improvement along the trajectory. We identified the location of the circle using the area of improvement heuristic described in Section~\ref{sec:automated-assessment}. In the event of a crash, we placed the circle on the location where the quadrotor crashed. If the quadrotor landed unsafely, we placed the circle at the point in the last 50 steps in the trajectory that had the worst robustness for landing speed or landing angle. For a safe landing, we placed the circle at the point in the trajectory with the highest combined control inputs.

\begin{figure*}[ht]
\centering
\begin{subfigure}[b]{\linewidth}
    \centering
   \includegraphics[width=.8\linewidth]{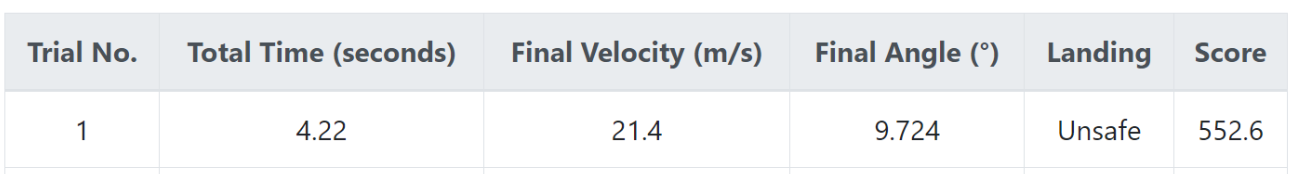}
   \caption{Baseline feedback condition}
   \label{fig:score-feedback}
   \vspace{10px}
\end{subfigure}
\begin{subfigure}[b]{\linewidth}
   \includegraphics[width=\linewidth]{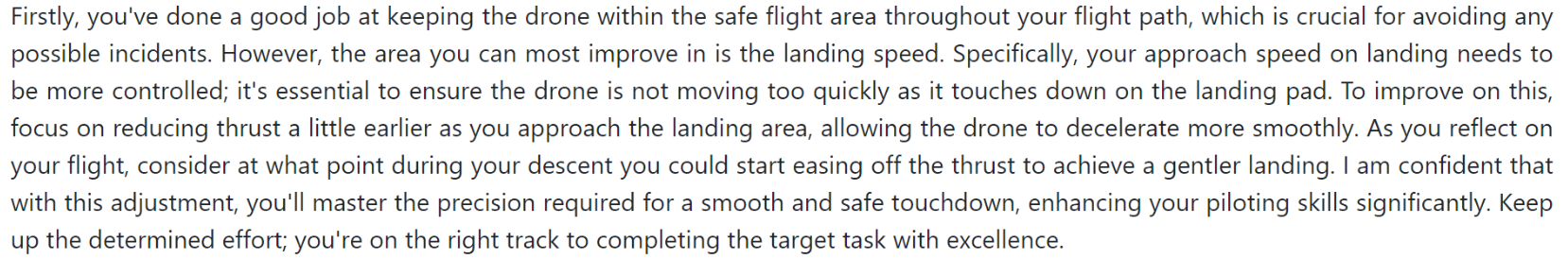}
   \caption{Text feedback condition}
   \label{fig:text-feedback}
   \vspace{10px}
\end{subfigure}

\begin{subfigure}[b]{.87\linewidth}
   \includegraphics[width=\linewidth]{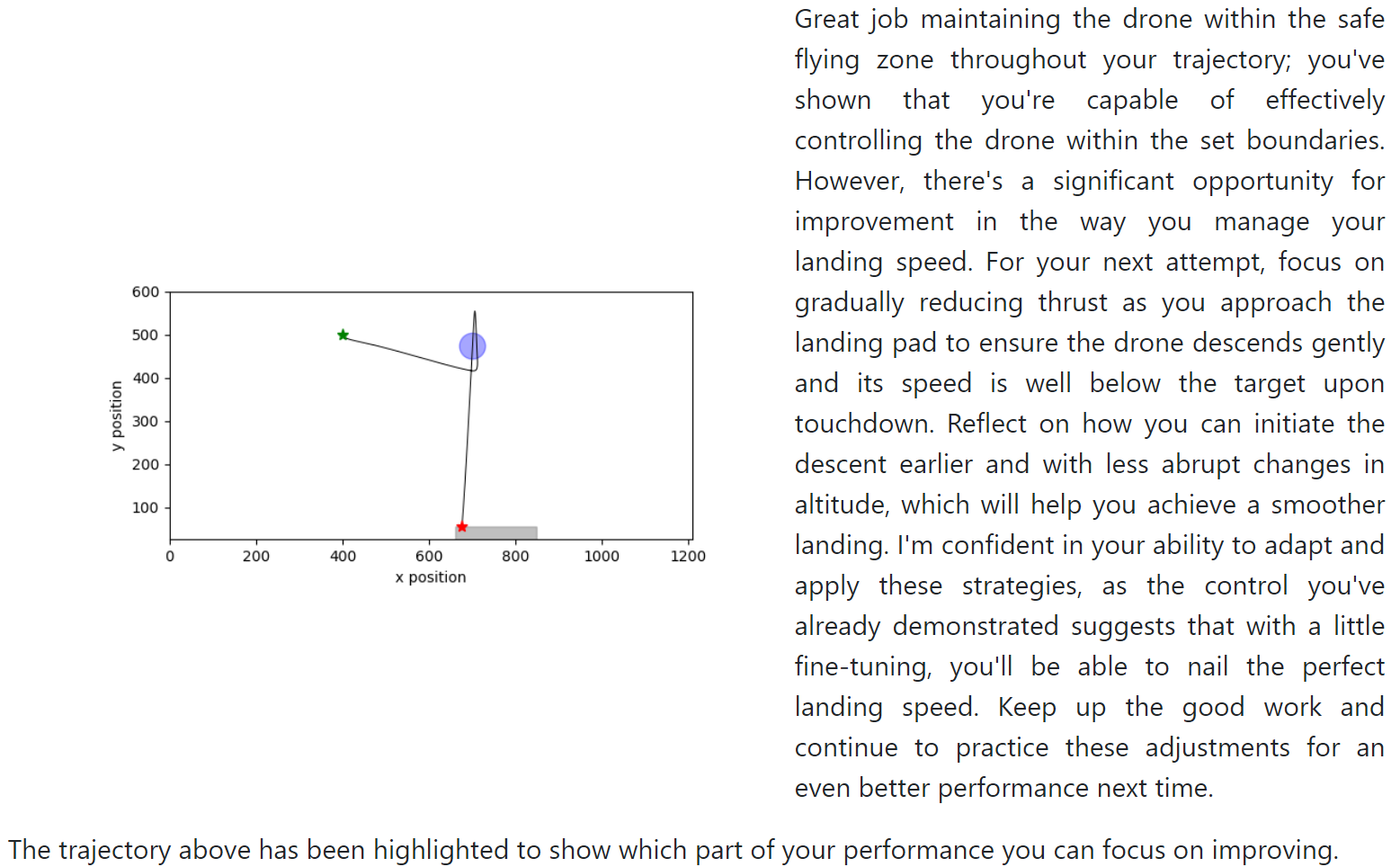}
   \caption{Multimodal feedback condition}
   \label{fig:full-feedback}
\end{subfigure}
\Description{The first image is a table showing the baseline feedback condition. The second image is a paragraph of text showing the text feedback condition. The third image has a graph of a trajectory on the left and a paragraph of text on the right, showing the multimodal feedback condition.}
\caption{Examples of the three feedback conditions used in the experiment.}
\label{fig:feedback-examples}
\end{figure*}

Figure~\ref{fig:feedback-examples} shows an example of each type of feedback. We generated the full set of text and image feedback regardless of condition so participants waited the same amount of time between trials.

\subsection{Measures}

\textit{Subjective Measures.} After each trial, participants rated the feedback they received. The purpose of the survey items was to understand how the generated feedback aligned with the desired dimensions of formative feedback described in Section~\ref{sec:rw-formative-feedback}. Table~\ref{tab:feedback-dimensions} summarizes the survey items participants completed after each trial. After completing the experiment, participants completed an exit survey that recorded their gender identity, age, experience flying drones, and video game experience. Participants also rated how helpful the feedback was overall (``The feedback I received helped me perform better on the task''; 1 = Strongly disagree, 5 = Strongly agree) and provided a text response discussing how the feedback influenced their piloting strategy over time.

\textit{Objective Measures.} We recorded the trajectory for each trial. The trajectory data included the quadrotor's $x$ and $y$ position and velocity, the quadrotor's rotation, and the participant's control throttle and attitude inputs. For each time step, we also calculated the trajectory's robustness according to the specifications in Section~\ref{sec:automated-assessment}. We recorded both trajectory and robustness data at 50 Hz.

\begin{table*}[htb]
\caption{Summary of the survey items participants completed after receiving feedback for a given trial. Participants rated their feedback using these items for each of the 20 trials in the experiment.}
\label{tab:feedback-dimensions}
\centering
\begin{tabular}{lll}
\toprule
Feedback Dimension & Survey Item                                                & Response Options                            \\
\midrule
Motivation         & ``The feedback motivated me to do better in future trials.'' & 1 = Strongly disagree, 5 = Strongly agree   \\
Manageable         & ``How much information did the feedback give?''              & 1 = Much too little, 5 = Much too much      \\
Actionable         & ``The feedback suggestions were actionable.''                & 1 = Strongly disagree, 5 = Strongly agree   \\
Timely             & ``How often was the feedback presented?''                    & 1 = Much too infrequent, 5 = Much too often \\
Reflection         & ``The feedback prompted me to reflect on my performance.''   & 1 = Strongly disagree, 5 = Strongly agree \\
\bottomrule
\end{tabular}
\end{table*}

\subsection{Data Analysis}

RQs 1 and 2 ask how participants perceived the feedback they received. The variables of interest were the subjective measures on each feedback dimension shown in Table~\ref{tab:feedback-dimensions} and the overall rating of feedback helpfulness, which yielded ordinal values. We found little discrimination between the extreme values of the Likert scales (Strongly Agree vs. Agree and Strongly Disagree vs. Disagree) so we collapsed these measures to a three-point scale (Disagree, Neutral, Agree) for analysis.

To answer RQ 1, we used the Kruskal-Wallis H-test to test for differences in feedback ratings between groups. As mentioned above, participants rated each dimension of feedback after every trial. To create independent samples, we aggregated survey responses for each participant across trials by calculating the most common response for each item. We found that participant ratings do not change much over time, which suggests that this method of aggregation provides an overall rating of each dimension of feedback.

We used ordered logistic regression models to answer RQ 2. The outcome variables were each participant's overall rating for each feedback dimension and their overall rating of the feedback's helpfulness. The independent variables included participant demographics, total number of safe landings, average trial time, and average time spent reviewing feedback. We also performed a trial-wise analysis of the feedback ratings, using trial time, feedback time, trial number, and landing outcome as predictors. The coefficients of these models ($\beta$) represent log-odds; we also report odds-ratios as $OR$ to aid with interpretation.

For RQ3, we considered several metrics of learning trajectory. We first evaluated mastery of the quadrotor landing task by calculating how many participants in each condition achieved at least one safe landing across the 20 trials. We used Fisher's Exact Test to test for differences between feedback conditions. We also considered how much participants improved in the task over time. We measured this by calculating how many more safe landings each participant achieved in the second half of the trials compared to the first half. To compare differences between feedback conditions, we used an independent-samples t-test.

\section{Results}
RQ 1 asks if participants in different feedback conditions perceived the dimensions of formative feedback differently. There is a statistically significant difference in ratings along the manageable ($H(2)=18.0,~p<0.001$) and actionable ($H(2)=18.1,~ p<0.001$) dimensions. Post-hoc Dunn's test with Bonferroni corrections reveals a significant difference in ratings along the manageable dimension between the baseline and text feedback conditions ($p<0.001$) and between the baseline and multimodal feedback conditions ($p=0.005$). There is also a significant difference in ratings along the actionable dimension between the baseline and text feedback conditions ($p<0.001$) and between the baseline and multimodal feedback conditions ($p<0.001$). There are no differences in ratings between the non-baseline feedback conditions. There are no significant differences in ratings for the motivation, timely, or reflection dimensions between the feedback groups.

\begin{figure}[ht]
\centering
\begin{subfigure}[b]{\columnwidth}
    \centering
   \includegraphics[width=\linewidth]{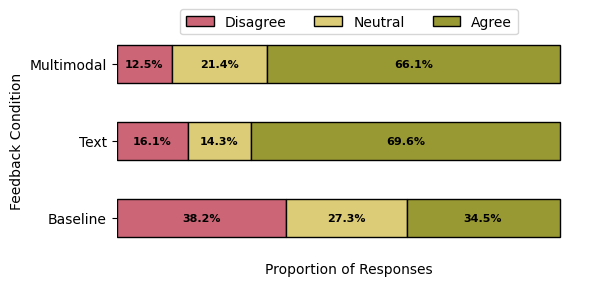}
   \caption{Distribution of responses for actionable feedback dimension: ``The feedback suggestions were actionable.''}
   \label{fig:actionable-likert}
   \vspace{10px}
\end{subfigure}
\begin{subfigure}[b]{\columnwidth}
   \includegraphics[width=\linewidth]{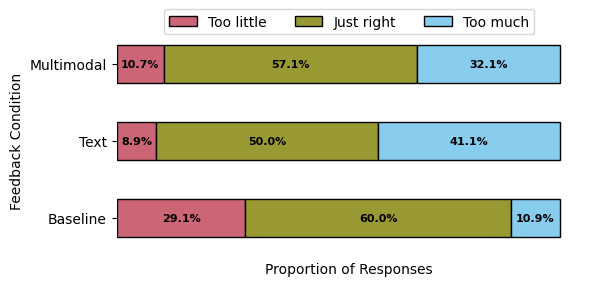}
   \caption{Distribution of responses for manageable feedback dimension: ``How much information did the feedback give?''}
   \label{fig:manageable-likert}
   \vspace{10px}
\end{subfigure}
\Description{Both images are stacked horizontal bar charts showing the distribution of likert responses for each condition. The top image is red, yellow, and green. The bottom image is red, green, and blue.}
\caption{Distributions of responses for feedback dimensions that were significantly different between groups. We show the responses collapsed to a three-point likert scale.}
\label{fig:likert-scales}
\end{figure}

A closer investigation of the distributions of survey responses shows that while participants in all conditions rate the manageability of the feedback as the right amount of information (50-60\% of participants in each condition), more participants rate the generated feedback conditions as providing too much information (32-41\%). Participants in the baseline condition are more likely to rate the feedback as having not enough information (29\%). More participants receiving generated feedback agree that the feedback was actionable (66-70\% of participants), while only 35\% of participants in the baseline condition agree that the feedback was actionable. Figure~\ref{fig:likert-scales} shows the distributions of ratings for the manageable and actionable feedback dimensions.

Although there are no significant differences between groups, participants as a whole generally find the feedback to be motivational (58-64\% agree) and prompting reflection (64-70\% agree). Participants also report similar ratings for the timely dimension, with 25-39\% reporting that feedback was delivered too often.

There is no significant difference between groups regarding where they found the feedback helpful to improving their performance. The majority of participants in the generated feedback conditions agree that the feedback benefited their performance (61-66\%) and 45\% of participants in the baseline condition agree that the feedback helped their performance on the task.

RQ 2 asked what factors predict the perception of formative feedback. We first fit an ordered logistic regression model to predict the rating of whether the feedback helped improve performance. While none of the demographic variables are significant predictors, four of the five aggregate measures of formative feedback are significant ($p<0.05$). Participants with higher `motivation' ($\beta = +0.97,~OR = 2.36$) and higher `reflection' ($\beta = +0.76,~OR = 2.14$) responses are more likely to rate the feedback as more helpful. Participants with higher `timely' responses are more likely to rate the feedback as less helpful ($\beta = -1.12,~OR = 0.33$), since higher `timely' ratings correspond to the perception that the feedback was given too often. Those with higher `manageable' ratings rate the feedback as more helpful ($\beta=+1.10,~OR=3.01$), which is interesting because higher manageable ratings mean the feedback contained too much information.

There are few variables that predict overall ratings for the elements of formative feedback. Participants with more experience with drones were more likely to rate the manageable dimension as having not enough information ($\beta=-0.80,~OR=0.45$). Those who achieved more safe landings are more likely to rate the feedback as occurring too often ($\beta=+0.09,~OR=1.10$). Older participants rate the feedback as promoting more reflection ($\beta=+0.06,~OR=1.06$). Finally, participants that spent more time completing the experiment rate the feedback as promoting less reflection, although this difference is small ($\beta=-0.001,~OR=1.00$).

Table~\ref{tab:trial-wise-predictors} summarizes which variables predicted feedback ratings at the trial level. Motivation, reflection, and actionable ratings increase with longer trial times and more time spent reviewing feedback. Participants view feedback more negatively as the experiment progresses, with higher trial numbers corresponding to lower motivation, reflection, and actionable ratings. Participants also rate the feedback as containing too much information and occurring too often when they achieve more successful landings.

\begin{table*}[ht]
\caption{Predictors of trial-wise feedback ratings for each dimension. We report only significant ($p<0.05$) coefficients as $\beta$ with corresponding odds-ratios.}
\label{tab:trial-wise-predictors}
\centering
\begin{tabular}{lccccc}
\toprule
Predictor & Motivation ($\beta$, OR) & Manageable ($\beta$, OR) & Timely ($\beta$, OR) & Reflection ($\beta$, OR) & Actionable ($\beta$, OR) \\
\midrule
Trial Time (s) & +0.005, 1.00 & & -0.01, 0.99 & +0.01, 1.01 & +0.004, 1.00 \\
Feedback Time (s) & +0.01, 1.01 & & +0.01, 1.01 & +0.01, 1.01 & +0.01, 1.01 \\
Trial Number & -0.03, 0.97 & & +0.02, 1.02 & -0.03, 0.97 & -0.02, 0.98 \\
Type of Landing & & +0.15, 1.17 & +0.33, 1.40 & & \\
\bottomrule
\end{tabular}
\end{table*}

RQ 3 asks how learning trajectories differed between groups. Fisher's Exact Test shows a significant difference in the number of people who failed to achieve a safe landing in any of the trials between the multimodal feedback condition and the baseline condition ($p = 0.04$). Only two participants fail to achieve a safe landing in the multimodal feedback condition (3.6\%), compared to eight participants in the baseline condition (14.5\%). There are no differences between the multimodal and text feedback conditions or the text and baseline feedback conditions.

All groups improved their performance at the task, as demonstrated by fewer crashes and more safe landings in the second half of the trials (see Table~\ref{tab:landings}). Participants in the multimodal feedback condition show a larger increase in safe landings ($M=2.4$ more safe landings, $SD=2.3$) compared to the text feedback condition ($M=1.4$, $SD=2.7$), $t(110) = 2.2$, $p = 0.03$. There are no significant differences between the baseline condition and the other feedback conditions.

\begin{table*}[ht]
\centering
\caption{Average (SD) number of landings for each feedback condition, calculated for the first and second half of the trials and for the whole experiment. Improvement is the average (SD) number more landings in the second half of trials.}
\label{tab:landings}
\begin{tabular}{lllll|llll}
\toprule
           & \multicolumn{4}{c|}{Safe Landings}                      & \multicolumn{4}{c}{Safe and Unsafe Landings}            \\
Condition  & Trials 1-10 & Trials 11-20 & Improvement & All Trials  & Trials 1-10 & Trials 11-20 & Improvement & All Trials   \\
\midrule
Baseline   & 3.42 (2.85) & 5.40 (3.28)  & 1.98 (2.18) & 8.82 (5.75) & 7.04 (2.24) & 8.53 (1.93)  & 1.49 (1.92) & 15.56 (3.71) \\
Text       & 3.05 (2.96) & 4.41 (3.07)  & 1.36 (2.65) & 7.46 (5.42) & 6.55 (2.61) & 8.20 (2.02)  & 1.64 (2.19) & 14.75 (4.17) \\
Multimodal & 2.75 (2.46) & 5.12 (2.52)  & 2.38 (2.33) & 7.88 (4.39) & 6.93 (2.21) & 8.89 (1.27)  & 1.96 (1.93) & 15.82 (3.06) \\
\bottomrule
\end{tabular}
\end{table*}

\section{Discussion}

\subsection{Main Findings}

In this paper, we developed an end-to-end training system that assesses performance and provides actionable feedback on a quadrotor landing task with no human intervention. The system uses temporal logic task specifications and task demonstrations to assess performance and provide formative feedback to the learner. We found significant differences in how manageable and actionable participants in the different conditions perceived the feedback. Importantly, we found several differences in learning outcomes between conditions. \textbf{Participants receiving multimodal feedback were more likely to safely land the quadrotor and showed greater improvement in safe landings in the second half of trials} compared to other feedback conditions. 

Overall, participants in all conditions had favorable views of the feedback they received. In particular, participants in the formative feedback conditions mentioned how the feedback impacted their motivation and self-confidence. One participant in the text feedback condition noted, ``Overall the encouragement was genuinely nice to receive, and helped to give motivation in completing the task and wanting to do well.'' Participants in the text and multimodal feedback conditions also reported the feedback felt personalized to their own skills and struggles. Another participant receiving text feedback reported, ``The feedback actually felt tailored to me, and not just the same stock answer every time.''

Participants in the baseline condition showed surprisingly positive perceptions of their feedback. The written responses indicated that participants were motivated by wanting to figure out how to improve their performance score. One participant noted, ``I tried to tell which criteria affected the score more, and how.'' Although we did not specifically design the baseline condition to be motivational and engaging, this result is in line with work showing that feedback can be intrinsic to the learner \cite{butler_feedback_1995}. Future works can investigate how to integrate feedback with principles of self-regulated and gamified learning. 

However, we found that participants naturally differentiated between performance data and formative feedback. In particular, participants in the baseline condition pushed back against labelling the data summary as feedback. Participants in this condition noted, ``The feedback did not help much with  strategizing, but it did make me want to get better scores.'' and ``The feedback didn't seem like feedback, because there was no suggestions. The feedback ... was just the numbers that we scored.'' These findings show that both the content and the delivery of feedback matters. Many of the studies discussed in Section~\ref{sec:related-work} implemented feedback similar to our baseline condition in the form of summary metrics. This feedback may be effective by providing learners with more information about their performance, but this depends on the learner to be able to interpret and devise new strategies based on their data. Truly formative feedback should help the learner interpret their data and act on it in future practice.

Participants receiving multimodal feedback were more likely to achieve a safe landing compared to the baseline condition. This could be due to how the multimodal feedback was personalized to address the learner's greatest area of improvement, while the summary statistics in the baseline condition remained the same regardless of performance. For example, one participant in the multimodal feedback condition said, ``The feedback helped tremendously by showing the exact location the unnecessary movements where at.'' Additionally, the format of the formative feedback may have encouraged participants to experiment with new control strategies. Another participant noted, ``The feedback helped me feel more confident in the adjustments I was making and to try new approaches.''

Finally, participants in the multimodal feedback condition improved more than the text feedback condition by achieving more safe landings in the second half of the experiment compared to the first half. This may be due to additional information provided by the annotated trajectory in the multimodal feedback condition. With the annotated trajectory, participants can pair general ideas presented in the text with a concrete emphasis on a particular area highlighted on the trajectory. In both conditions, participants noted that the feedback did not give specific enough strategies to improve performance. One participant in the text feedback condition noted, ``It asked me to consider how changing it ``might'' be more effective but not exactly how (try using the W key more often to keep the drone up longer, for example).''

These observations highlight an area to improve the feedback prompt template. When designing the feedback, we prompted the model to use actionable suggestions related to the throttle and rotation of the quadrotor. While these terms are specific to quadrotors and other aircraft, they did not tell the learner exactly what to do (e.g. what buttons to push and how) to perform better in this particular simulation environment. This suggests feedback can be actionable on several levels, depending on the complexity of the task one is learning.

\subsection{Emerging Themes}
The results from this study illuminate a need to consider how to adapt feedback beyond the most recent trial. For example, the approach presented here does not consider persistent skill gaps that appear over several trials. One participant in the text feedback condition wrote, ``I wish the feedback generated built on the performance in prior trials so the feedback could say \textit{you've improved!} instead of \textit{you need to be better at the same thing... even though you actually did improve compared to the last trial}.''

Additionally, we can use different feedback strategies depending on the overall task performance; high-performing individuals may only need to reinforce their successful control strategies while novices may need more structured and specific feedback presented in this study. Several participants noted frustration when receiving feedback after a successful landing. One participant in the text feedback condition noted, ``It is a little discouraging to finally make a successful landing, and then get \textit{a yeah, you did it - but you should focus on doing it better}.''

Future work should also consider how to schedule feedback over time. Several participants reported ignoring the feedback as the trial progressed, especially if they were consistently performing well on the task. A participant receiving multimodal feedback wrote, ``After finding the fastest way of landing the drone, I did not follow any more suggestions.'' Additionally, Participants also reported needing time to independently explore the dynamics of the task before receiving performance feedback. A participant in the text feedback condition reported, ``It would benefit me to go straight into the next trial so I can continue to make small adjustments back to back... Half of learning is trial and error.'' 

Recent work discusses how prompt-based generative feedback methods are ideal for quickly prototyping and testing feedback templates \cite{jensen_2024_llms}. Future works can investigate using simple rules to determine what feedback template to generate. How to adjust the timing of formative feedback based on the number of attempts and performance outcomes remains an open question.

As automated training systems continue to develop, it is important to consider their place among other workplace programs. It is likely we will need to balance automated approaches with more traditional one-on-one training \cite{hutson_rethinking_2023}; in addition to learning technical skills, training programs will need to consider the social aspects of learning such as developing a community of practice within an organization \cite{leon_employees_2023}. As required workplace skills and knowledge continue to develop over time, training systems will need to both provide initial background knowledge and additional support to help workers remain up-to-date \cite{morandini_impact_2023}.

\subsection{Study Limitations}
This work is limited in several ways. First, the quadrotor landing task did not change between trials. This means that when participants found a control strategy that yielded a successful result, they tended to repeat the same strategy. Future works may wish to randomize the starting point of the drone in the simulation to provide more insight about if participants are learning strategies that transfer to other landing scenarios. 

The other main limitation of this study is the online nature of the data collection. While the Prolific platform allowed us to quickly recruit a large sample of participants, we were not able to observe nuanced reactions to the feedback they received. Future work can pair crowd-sourced methods with in-person studies to understand how participants choose to integrate feedback into their learning process.

\section{Conclusions}

In this paper, we developed an adaptive training system for a simulated quadrotor landing task. The system first assesses performance based on temporal logic specifications, which require no prior data to learn and can be flexibly adapted to new tasks and situations. Using these assessment results, we automatically generated multimodal feedback adhering to principles of effective formative feedback. While participants in all conditions reported finding the feedback engaging and motivating, they differed in their ratings of how actionable and manageable the feedback was. Since the goal of a training system is to help learners master a new task, we also considered learning differences between conditions. We found that participants receiving multimodal feedback were more likely to achieve a safe landing. They also improved more over the course of the experiment by increasing their safe landings more compared to other feedback conditions. Based on these results, we identified future opportunities to further adapt feedback over time and consider the learner's affective state when delivering feedback. While future work in psychomotor task training will continue to depend on domain-specific methods and knowledge, we encourage researchers to align their methods with established pedagogical theories of learning and feedback.

\begin{acks}
This work was supported by the US National Science Foundation (NSF) under award number 1836900.
\end{acks}

\bibliographystyle{ACM-Reference-Format}
\bibliography{refs}


\begin{thebibliography}{49}

%%% the Web address.

\ifx \showCODEN    \undefined \def \showCODEN     #1{\unskip}     \fi
\ifx \showDOI      \undefined \def \showDOI       #1{#1}\fi
\ifx \showISBNx    \undefined \def \showISBNx     #1{\unskip}     \fi
\ifx \showISBNxiii \undefined \def \showISBNxiii  #1{\unskip}     \fi
\ifx \showISSN     \undefined \def \showISSN      #1{\unskip}     \fi
\ifx \showLCCN     \undefined \def \showLCCN      #1{\unskip}     \fi
\ifx \shownote     \undefined \def \shownote      #1{#1}          \fi
\ifx \showarticletitle \undefined \def \showarticletitle #1{#1}   \fi
\ifx \showURL      \undefined \def \showURL       {\relax}        \fi
% invisible to TeX
\providecommand\bibfield[2]{#2}
\providecommand\bibinfo[2]{#2}
\providecommand\natexlab[1]{#1}
\providecommand\showeprint[2][]{arXiv:#2}

\bibitem[Andrew et~al\mbox{.}(2020)]%
        {andrew_analysis_2020}
\bibfield{author}{\bibinfo{person}{Megan Andrew}, \bibinfo{person}{Timothy Marler}, \bibinfo{person}{Jesse Lastunen}, \bibinfo{person}{Hannah Acheson-Field}, {and} \bibinfo{person}{Steven Popper}.} \bibinfo{year}{2020}\natexlab{}.
\newblock \bibinfo{booktitle}{\emph{An {Analysis} of {Education} and {Training} {Programs} in {Advanced} {Manufacturing} {Using} {Robotics}}}.
\newblock \bibinfo{publisher}{RAND Corporation}, \bibinfo{address}{Pittsburgh, Pennsylvania}.
\newblock
\urldef\tempurl%
\url{https://doi.org/10.7249/RR4244}
\showDOI{\tempurl}


\bibitem[Banihashem et~al\mbox{.}(2022)]%
        {banihashem_systematic_2022}
\bibfield{author}{\bibinfo{person}{Seyyed~Kazem Banihashem}, \bibinfo{person}{Omid Noroozi}, \bibinfo{person}{Stan Van~Ginkel}, \bibinfo{person}{Leah~P. Macfadyen}, {and} \bibinfo{person}{Harm~J.A. Biemans}.} \bibinfo{year}{2022}\natexlab{}.
\newblock \showarticletitle{A systematic review of the role of learning analytics in enhancing feedback practices in higher education}.
\newblock \bibinfo{journal}{\emph{Educational Research Review}}  \bibinfo{volume}{37} (\bibinfo{date}{Nov.} \bibinfo{year}{2022}), \bibinfo{pages}{100489}.
\newblock
\showISSN{1747938X}
\urldef\tempurl%
\url{https://doi.org/10.1016/j.edurev.2022.100489}
\showDOI{\tempurl}


\bibitem[Bell(1985)]%
        {bell_employment_1985}
\bibfield{author}{\bibinfo{person}{D.~A. Bell}.} \bibinfo{year}{1985}\natexlab{}.
\newblock \showarticletitle{Employment skills for the robot age}.
\newblock \bibinfo{journal}{\emph{Robotica}} \bibinfo{volume}{3}, \bibinfo{number}{2} (\bibinfo{date}{April} \bibinfo{year}{1985}), \bibinfo{pages}{93--95}.
\newblock
\showISSN{1469-8668, 0263-5747}
\urldef\tempurl%
\url{https://doi.org/10.1017/S0263574700001788}
\showDOI{\tempurl}


\bibitem[Buche et~al\mbox{.}(2003)]%
        {buche_mascaret_2003}
\bibfield{author}{\bibinfo{person}{C. Buche}, \bibinfo{person}{R. Querrec}, \bibinfo{person}{P. De~Loor}, {and} \bibinfo{person}{P. Chevaillier}.} \bibinfo{year}{2003}\natexlab{}.
\newblock \showarticletitle{{MASCARET}: pedagogical multi-agents systems for virtual environment for training}. In \bibinfo{booktitle}{\emph{Proceedings. 2003 {International} {Conference} on {Cyberworlds}}}. \bibinfo{publisher}{IEEE Comput. Soc}, \bibinfo{address}{Singapore}, \bibinfo{pages}{423--430}.
\newblock
\showISBNx{978-0-7695-1922-7}
\urldef\tempurl%
\url{https://doi.org/10.1109/CYBER.2003.1253485}
\showDOI{\tempurl}


\bibitem[Butler and Winne(1995)]%
        {butler_feedback_1995}
\bibfield{author}{\bibinfo{person}{Deborah~L. Butler} {and} \bibinfo{person}{Philip~H. Winne}.} \bibinfo{year}{1995}\natexlab{}.
\newblock \showarticletitle{Feedback and self-regulated learning: {A} theoretical synthesis}.
\newblock \bibinfo{journal}{\emph{Review of Educational Research}} \bibinfo{volume}{65}, \bibinfo{number}{3} (\bibinfo{year}{1995}), \bibinfo{pages}{245--281}.
\newblock
\showISSN{1935-1046}
\urldef\tempurl%
\url{https://doi.org/10.2307/1170684}
\showDOI{\tempurl}
\newblock
\shownote{Place: US Publisher: American Educational Research Assn}.


\bibitem[Davaris et~al\mbox{.}(2019)]%
        {davaris_importance_2019}
\bibfield{author}{\bibinfo{person}{Myles Davaris}, \bibinfo{person}{Sudanthi Wijewickrema}, \bibinfo{person}{Yun Zhou}, \bibinfo{person}{Patorn Piromchai}, \bibinfo{person}{James Bailey}, \bibinfo{person}{Gregor Kennedy}, {and} \bibinfo{person}{Stephen O'Leary}.} \bibinfo{year}{2019}\natexlab{}.
\newblock \showarticletitle{The {Importance} of {Automated} {Real}-{Time} {Performance} {Feedback} in {Virtual} {Reality} {Temporal} {Bone} {Surgery} {Training}}. In \bibinfo{booktitle}{\emph{Proceedings of the 2019 Artificial Intelligence in Education Conference}}, \bibfield{editor}{\bibinfo{person}{Seiji Isotani}, \bibinfo{person}{Eva Millán}, \bibinfo{person}{Amy Ogan}, \bibinfo{person}{Peter Hastings}, \bibinfo{person}{Bruce McLaren}, {and} \bibinfo{person}{Rose Luckin}} (Eds.). \bibinfo{publisher}{Springer International Publishing}, \bibinfo{address}{Cham}, \bibinfo{pages}{96--109}.
\newblock
\showISBNx{978-3-030-23204-7}


\bibitem[Di~Mitri et~al\mbox{.}(2022)]%
        {di_mitri_keep_2022}
\bibfield{author}{\bibinfo{person}{Daniele Di~Mitri}, \bibinfo{person}{Jan Schneider}, {and} \bibinfo{person}{Hendrik Drachsler}.} \bibinfo{year}{2022}\natexlab{}.
\newblock \showarticletitle{Keep {Me} in the {Loop}: {Real}-{Time} {Feedback} with {Multimodal} {Data}}.
\newblock \bibinfo{journal}{\emph{Int J Artif Intell Educ}} \bibinfo{volume}{32}, \bibinfo{number}{4} (\bibinfo{date}{Dec.} \bibinfo{year}{2022}), \bibinfo{pages}{1093--1118}.
\newblock
\showISSN{1560-4292, 1560-4306}
\urldef\tempurl%
\url{https://doi.org/10.1007/s40593-021-00281-z}
\showDOI{\tempurl}


\bibitem[Donz{\'e} and Maler(2010)]%
        {Donze+Maler/2010/Robust}
\bibfield{author}{\bibinfo{person}{Alexandre Donz{\'e}} {and} \bibinfo{person}{Oded Maler}.} \bibinfo{year}{2010}\natexlab{}.
\newblock \showarticletitle{Robust Satisfaction of Temporal Logic over Real-Valued Signals}.
\newblock In \bibinfo{booktitle}{\emph{FORMATS}}. \bibinfo{series}{Lecture Notes in Computer Science}, Vol.~\bibinfo{volume}{6246}. \bibinfo{publisher}{Springer}, \bibinfo{pages}{92--106}.
\newblock


\bibitem[Echeverria and C.~Santos(2021)]%
        {echeverria_2021_ws}
\bibfield{author}{\bibinfo{person}{Jon Echeverria} {and} \bibinfo{person}{Olga C.~Santos}.} \bibinfo{year}{2021}\natexlab{}.
\newblock \showarticletitle{KUMITRON: Artificial Intelligence System to Monitor Karate Fights that Synchronize Aerial Images with Physiological and Inertial Signals}. In \bibinfo{booktitle}{\emph{Companion Proceedings of the 26th International Conference on Intelligent User Interfaces}} (, College Station, TX, USA,) \emph{(\bibinfo{series}{IUI '21 Companion})}. \bibinfo{publisher}{Association for Computing Machinery}, \bibinfo{address}{New York, NY, USA}, \bibinfo{pages}{37–39}.
\newblock
\showISBNx{9781450380188}
\urldef\tempurl%
\url{https://doi.org/10.1145/3397482.3450730}
\showDOI{\tempurl}


\bibitem[Echeverria and Santos(2021)]%
        {echeverria_2021}
\bibfield{author}{\bibinfo{person}{Jon Echeverria} {and} \bibinfo{person}{Olga~C. Santos}.} \bibinfo{year}{2021}\natexlab{}.
\newblock \showarticletitle{Toward Modeling Psychomotor Performance in Karate Combats Using Computer Vision Pose Estimation}.
\newblock \bibinfo{journal}{\emph{Sensors}} \bibinfo{volume}{21}, \bibinfo{number}{24} (\bibinfo{year}{2021}), \bibinfo{numpages}{27}~pages.
\newblock
\showISSN{1424-8220}
\urldef\tempurl%
\url{https://doi.org/10.3390/s21248378}
\showDOI{\tempurl}


\bibitem[Gregory and Ward(2000)]%
        {gregory_work_2000}
\bibfield{author}{\bibinfo{person}{Rita~A. Gregory} {and} \bibinfo{person}{Terrance Ward}.} \bibinfo{year}{2000}\natexlab{}.
\newblock \showarticletitle{Work {Force} {Characteristics} in a {Robot} {Driven} {Construction} {Industry}}. In \bibinfo{booktitle}{\emph{Proceedings of the 17th {IAARC}/{CIB}/{IEEE}/{IFAC}/{IFR} {International} {Symposium} on {Automation} and {Robotics} in {Construction}}}. \bibinfo{publisher}{International Association for Automation and Robotics in Construction (IAARC)}, \bibinfo{address}{Taipei, Taiwan}, \bibinfo{pages}{1--5}.
\newblock
\showISBNx{978-957-02-6698-6}
\urldef\tempurl%
\url{https://doi.org/10.22260/ISARC2000/0072}
\showDOI{\tempurl}
\newblock
\shownote{ISSN: 2413-5844}.


\bibitem[Haslgr\"{u}bler et~al\mbox{.}(2019)]%
        {haslgrubler_2019}
\bibfield{author}{\bibinfo{person}{Michael Haslgr\"{u}bler}, \bibinfo{person}{Benedikt Gollan}, \bibinfo{person}{Christian Thomay}, \bibinfo{person}{Alois Ferscha}, {and} \bibinfo{person}{Josef Heftberger}.} \bibinfo{year}{2019}\natexlab{}.
\newblock \showarticletitle{Towards skill recognition using eye-hand coordination in industrial production}. In \bibinfo{booktitle}{\emph{Proceedings of the 12th ACM International Conference on PErvasive Technologies Related to Assistive Environments}} (Rhodes, Greece) \emph{(\bibinfo{series}{PETRA '19})}. \bibinfo{publisher}{Association for Computing Machinery}, \bibinfo{address}{New York, NY, USA}, \bibinfo{pages}{11–20}.
\newblock
\showISBNx{9781450362320}
\urldef\tempurl%
\url{https://doi.org/10.1145/3316782.3316784}
\showDOI{\tempurl}


\bibitem[Hatziapostolou and Paraskakis(2010)]%
        {hatziapostolou_enhancing_2010}
\bibfield{author}{\bibinfo{person}{Thanos Hatziapostolou} {and} \bibinfo{person}{Iraklis Paraskakis}.} \bibinfo{year}{2010}\natexlab{}.
\newblock \showarticletitle{Enhancing the {Impact} of {Formative} {Feedback} on {Student} {Learning} {Through} an {Online} {Feedback} {System}}.
\newblock \bibinfo{journal}{\emph{Electronic Journal of e-Learning}} \bibinfo{volume}{8}, \bibinfo{number}{2} (\bibinfo{year}{2010}), \bibinfo{pages}{111 -- 122}.
\newblock


\bibitem[Henderson et~al\mbox{.}(2019)]%
        {henderson_conditions_2019}
\bibfield{author}{\bibinfo{person}{Michael Henderson}, \bibinfo{person}{Michael Phillips}, \bibinfo{person}{Tracii Ryan}, \bibinfo{person}{David Boud}, \bibinfo{person}{Phillip Dawson}, \bibinfo{person}{Elizabeth Molloy}, {and} \bibinfo{person}{Paige Mahoney}.} \bibinfo{year}{2019}\natexlab{}.
\newblock \showarticletitle{Conditions that enable effective feedback}.
\newblock \bibinfo{journal}{\emph{Higher Education Research \& Development}} \bibinfo{volume}{38}, \bibinfo{number}{7} (\bibinfo{date}{Nov.} \bibinfo{year}{2019}), \bibinfo{pages}{1401--1416}.
\newblock
\showISSN{0729-4360, 1469-8366}
\urldef\tempurl%
\url{https://doi.org/10.1080/07294360.2019.1657807}
\showDOI{\tempurl}


\bibitem[Hutson and Ceballos(2023)]%
        {hutson_rethinking_2023}
\bibfield{author}{\bibinfo{person}{James Hutson} {and} \bibinfo{person}{Jason Ceballos}.} \bibinfo{year}{2023}\natexlab{}.
\newblock \showarticletitle{Rethinking {Education} in the {Age} of {AI}: {The} {Importance} of {Developing} {Durable} {Skills} in the {Industry} 4.0}.
\newblock \bibinfo{journal}{\emph{Journal of Information Economics}} \bibinfo{volume}{1}, \bibinfo{number}{2} (\bibinfo{date}{July} \bibinfo{year}{2023}), \bibinfo{pages}{26--35}.
\newblock
\urldef\tempurl%
\url{https://doi.org/10.58567/jie01020002}
\showDOI{\tempurl}


\bibitem[Jensen et~al\mbox{.}(2023)]%
        {jensen_more_2023}
\bibfield{author}{\bibinfo{person}{Emily Jensen}, \bibinfo{person}{Bradley Hayes}, {and} \bibinfo{person}{Sriram Sankaranarayanan}.} \bibinfo{year}{2023}\natexlab{}.
\newblock \showarticletitle{More {Than} a {Number}: {A} {Multi}-dimensional {Framework} {For} {Automatically} {Assessing} {Human} {Teleoperation} {Skill}}. In \bibinfo{booktitle}{\emph{Companion of the 2023 {ACM}/{IEEE} {International} {Conference} on {Human}-{Robot} {Interaction}}}. \bibinfo{publisher}{ACM}, \bibinfo{address}{Stockholm Sweden}, \bibinfo{pages}{653--657}.
\newblock
\showISBNx{978-1-4503-9970-8}
\urldef\tempurl%
\url{https://doi.org/10.1145/3568294.3580167}
\showDOI{\tempurl}


\bibitem[Jensen et~al\mbox{.}(2024)]%
        {jensen_2024_llms}
\bibfield{author}{\bibinfo{person}{Emily Jensen}, \bibinfo{person}{Sriram Sankaranarayanan}, {and} \bibinfo{person}{Bradley Hayes}.} \bibinfo{year}{2024}\natexlab{}.
\newblock \showarticletitle{Large Language Models Enable Automated Formative Feedback in Human-Robot Interaction Tasks}. In \bibinfo{booktitle}{\emph{Human – Large Language Model Interaction Workshop}}. \bibinfo{publisher}{ACM/IEEE}, \bibinfo{address}{Boulder, CO}, \bibinfo{numpages}{2}~pages.
\newblock


\bibitem[Korhonen et~al\mbox{.}(2023)]%
        {korhonen_training_2023}
\bibfield{author}{\bibinfo{person}{Tiina Korhonen}, \bibinfo{person}{Timo Lindqvist}, \bibinfo{person}{Joakim Laine}, {and} \bibinfo{person}{Kai Hakkarainen}.} \bibinfo{year}{2023}\natexlab{}.
\newblock \showarticletitle{Training {Hard} {Skills} in {Virtual} {Reality}: {Developing} a {Theoretical} {Framework} for {AI}-{Based} {Immersive} {Learning}}.
\newblock In \bibinfo{booktitle}{\emph{{AI} in {Learning}: {Designing} the {Future}}}, \bibfield{editor}{\bibinfo{person}{Hannele Niemi}, \bibinfo{person}{Roy~D. Pea}, {and} \bibinfo{person}{Yu~Lu}} (Eds.). \bibinfo{publisher}{Springer International Publishing}, \bibinfo{address}{Cham}, \bibinfo{pages}{195--213}.
\newblock
\showISBNx{978-3-031-09687-7}
\urldef\tempurl%
\url{https://doi.org/10.1007/978-3-031-09687-7_12}
\showDOI{\tempurl}


\bibitem[Laine et~al\mbox{.}(2022)]%
        {laine_systematic_2022}
\bibfield{author}{\bibinfo{person}{Joakim Laine}, \bibinfo{person}{Timo Lindqvist}, \bibinfo{person}{Tiina Korhonen}, {and} \bibinfo{person}{Kai Hakkarainen}.} \bibinfo{year}{2022}\natexlab{}.
\newblock \showarticletitle{Systematic {Review} of {Intelligent} {Tutoring} {Systems} for {Hard} {Skills} {Training} in {Virtual} {Reality} {Environments}}.
\newblock \bibinfo{journal}{\emph{International Journal of Technology in Education and Science}} \bibinfo{volume}{6}, \bibinfo{number}{2} (\bibinfo{date}{May} \bibinfo{year}{2022}), \bibinfo{pages}{178--203}.
\newblock
\showISSN{2651-5369}
\urldef\tempurl%
\url{https://doi.org/10.46328/ijtes.348}
\showDOI{\tempurl}
\newblock
\shownote{Number: 2}.


\bibitem[Leon(2023)]%
        {leon_employees_2023}
\bibfield{author}{\bibinfo{person}{Ramona~Diana Leon}.} \bibinfo{year}{2023}\natexlab{}.
\newblock \showarticletitle{Employees’ reskilling and upskilling for industry 5.0: {Selecting} the best professional development programmes}.
\newblock \bibinfo{journal}{\emph{Technology in Society}}  \bibinfo{volume}{75} (\bibinfo{date}{Nov.} \bibinfo{year}{2023}), \bibinfo{pages}{102393}.
\newblock
\showISSN{0160-791X}
\urldef\tempurl%
\url{https://doi.org/10.1016/j.techsoc.2023.102393}
\showDOI{\tempurl}


\bibitem[Li(2022)]%
        {li_reskilling_2022}
\bibfield{author}{\bibinfo{person}{Ling Li}.} \bibinfo{year}{2022}\natexlab{}.
\newblock \showarticletitle{Reskilling and {Upskilling} the {Future}-ready {Workforce} for {Industry} 4.0 and {Beyond}}.
\newblock \bibinfo{journal}{\emph{Inf Syst Front}}  \bibinfo{volume}{2022} (\bibinfo{date}{July} \bibinfo{year}{2022}), \bibinfo{numpages}{16}~pages.
\newblock
\showISSN{1572-9419}
\urldef\tempurl%
\url{https://doi.org/10.1007/s10796-022-10308-y}
\showDOI{\tempurl}


\bibitem[Luengo and Mufti-Alchawafa(2013)]%
        {luengo_target_2013}
\bibfield{author}{\bibinfo{person}{Vanda Luengo} {and} \bibinfo{person}{Dima Mufti-Alchawafa}.} \bibinfo{year}{2013}\natexlab{}.
\newblock \showarticletitle{Target the controls during the problem solving activity, a process to produce adapted epistemic feedbacks in ill- defined domains.}. In \bibinfo{booktitle}{\emph{Formative {Feedback} in {Interactive} {Learning} {Environments} ({FFILE}) {Conference}}}. \bibinfo{publisher}{Springer}, \bibinfo{address}{Memphis, USA}, \bibinfo{pages}{8}.
\newblock


\bibitem[Martinez et~al\mbox{.}(2023)]%
        {martinez_content-focused_2023}
\bibfield{author}{\bibinfo{person}{Cecilia Martinez}, \bibinfo{person}{Ramiro Serra}, \bibinfo{person}{Prem Sundaramoorthy}, \bibinfo{person}{Thomas Booij}, \bibinfo{person}{Cornelis Vertegaal}, \bibinfo{person}{Zahra Bounik}, \bibinfo{person}{Kevin Van~Hastenberg}, {and} \bibinfo{person}{Mark Bentum}.} \bibinfo{year}{2023}\natexlab{}.
\newblock \showarticletitle{Content-{Focused} {Formative} {Feedback} {Combining} {Achievement}, {Qualitative} and {Learning} {Analytics} {Data}}.
\newblock \bibinfo{journal}{\emph{Education Sciences}} \bibinfo{volume}{13}, \bibinfo{number}{10} (\bibinfo{date}{Oct.} \bibinfo{year}{2023}), \bibinfo{pages}{1014}.
\newblock
\showISSN{2227-7102}
\urldef\tempurl%
\url{https://doi.org/10.3390/educsci13101014}
\showDOI{\tempurl}


\bibitem[Mat~Sanusi et~al\mbox{.}(2021)]%
        {mat_sanusi_2021}
\bibfield{author}{\bibinfo{person}{Khaleel~Asyraaf Mat~Sanusi}, \bibinfo{person}{Daniele~Di Mitri}, \bibinfo{person}{Bibeg Limbu}, {and} \bibinfo{person}{Roland Klemke}.} \bibinfo{year}{2021}\natexlab{}.
\newblock \showarticletitle{Table Tennis Tutor: Forehand Strokes Classification Based on Multimodal Data and Neural Networks}.
\newblock \bibinfo{journal}{\emph{Sensors}} \bibinfo{volume}{21}, \bibinfo{number}{9} (\bibinfo{year}{2021}), \bibinfo{numpages}{18}~pages.
\newblock
\showISSN{1424-8220}
\urldef\tempurl%
\url{https://doi.org/10.3390/s21093121}
\showDOI{\tempurl}


\bibitem[Morandini et~al\mbox{.}(2023)]%
        {morandini_impact_2023}
\bibfield{author}{\bibinfo{person}{Sofia Morandini}, \bibinfo{person}{Federico Fraboni}, \bibinfo{person}{Marco De~Angelis}, \bibinfo{person}{Gabriele Puzzo}, \bibinfo{person}{Davide Diusino}, {and} \bibinfo{person}{Luca Pietrantoni}.} \bibinfo{year}{2023}\natexlab{}.
\newblock \showarticletitle{The {Impact} of {Artificial} {Intelligence} on {Workers}’ {Skills}: {Upskilling} and {Reskilling} in {Organisations}}.
\newblock \bibinfo{journal}{\emph{Informing Science: The International Journal of an Emerging Transdiscipline}}  \bibinfo{volume}{26} (\bibinfo{date}{Feb.} \bibinfo{year}{2023}), \bibinfo{pages}{039--068}.
\newblock
\urldef\tempurl%
\url{https://www.informingscience.org/Publications/5078}
\showURL{%
\tempurl}


\bibitem[Moringen et~al\mbox{.}(2021)]%
        {moringen2021optimizing}
\bibfield{author}{\bibinfo{person}{Alexandra Moringen}, \bibinfo{person}{Sören Rüttgers}, \bibinfo{person}{Luisa Zintgraf}, \bibinfo{person}{Jason Friedman}, {and} \bibinfo{person}{Helge Ritter}.} \bibinfo{year}{2021}\natexlab{}.
\newblock \bibinfo{title}{Optimizing piano practice with a utility-based scaffold}.
\newblock
\newblock
\showeprint[arxiv]{2106.12937}~[cs.HC]


\bibitem[Moya et~al\mbox{.}(2023)]%
        {moya_augmented_2023}
\bibfield{author}{\bibinfo{person}{Ana Moya}, \bibinfo{person}{Leire Bastida}, \bibinfo{person}{Pablo Aguirrezabal}, \bibinfo{person}{Matteo Pantano}, {and} \bibinfo{person}{Patricia Abril-Jiménez}.} \bibinfo{year}{2023}\natexlab{}.
\newblock \showarticletitle{Augmented {Reality} for {Supporting} {Workers} in {Human}–{Robot} {Collaboration}}.
\newblock \bibinfo{journal}{\emph{MTI}} \bibinfo{volume}{7}, \bibinfo{number}{4} (\bibinfo{date}{April} \bibinfo{year}{2023}), \bibinfo{pages}{40}.
\newblock
\showISSN{2414-4088}
\urldef\tempurl%
\url{https://doi.org/10.3390/mti7040040}
\showDOI{\tempurl}


\bibitem[Mukherjee et~al\mbox{.}(2023)]%
        {mukherjee_identification_2023}
\bibfield{author}{\bibinfo{person}{Abheek~Anjan Mukherjee}, \bibinfo{person}{Alok Raj}, {and} \bibinfo{person}{Shikha Aggarwal}.} \bibinfo{year}{2023}\natexlab{}.
\newblock \showarticletitle{Identification of barriers and their mitigation strategies for industry 5.0 implementation in emerging economies}.
\newblock \bibinfo{journal}{\emph{International Journal of Production Economics}}  \bibinfo{volume}{257} (\bibinfo{year}{2023}), \bibinfo{pages}{108770}.
\newblock
\showISSN{0925-5273}
\urldef\tempurl%
\url{https://doi.org/10.1016/j.ijpe.2023.108770}
\showDOI{\tempurl}


\bibitem[Nicholls et~al\mbox{.}(2014)]%
        {Nicholls2014}
\bibfield{author}{\bibinfo{person}{Delwyn Nicholls}, \bibinfo{person}{Linda Sweet}, {and} \bibinfo{person}{Jon Hyett}.} \bibinfo{year}{2014}\natexlab{}.
\newblock \showarticletitle{Psychomotor Skills in Medical Ultrasound Imaging}.
\newblock \bibinfo{journal}{\emph{Journal of Ultrasound in Medicine}} \bibinfo{volume}{33}, \bibinfo{number}{8} (\bibinfo{year}{2014}), \bibinfo{pages}{1349--1352}.
\newblock
\urldef\tempurl%
\url{https://doi.org/10.7863/ultra.33.8.1349}
\showDOI{\tempurl}
\showeprint{https://onlinelibrary.wiley.com/doi/pdf/10.7863/ultra.33.8.1349}


\bibitem[OpenAI(2023)]%
        {openai_gpt-4vision_2023}
\bibfield{author}{\bibinfo{person}{OpenAI}.} \bibinfo{year}{2023}\natexlab{}.
\newblock \bibinfo{title}{{GPT}-{4V}(ision) technical work and authors}.
\newblock
\newblock
\urldef\tempurl%
\url{https://openai.com/contributions/gpt-4v/}
\showURL{%
\tempurl}


\bibitem[Ortiz(2020)]%
        {ortiz_capturing_2020}
\bibfield{author}{\bibinfo{person}{Alberto~Casas Ortiz}.} \bibinfo{year}{2020}\natexlab{}.
\newblock \emph{\bibinfo{title}{Capturing, {Modelling}, {Analyzing} and providing {Feedback} in {Martial} {Arts} with {Artificial} {Intelligence} to support {Psychomotor} {Learning} {Activities}}}.
\newblock \bibinfo{thesistype}{Master's\ thesis}. \bibinfo{school}{Universidad Nacional de Educación a Distancia}.
\newblock


\bibitem[Pishchukhina and Allen(2021)]%
        {pishchukhina_supporting_2021}
\bibfield{author}{\bibinfo{person}{Olga Pishchukhina} {and} \bibinfo{person}{Angela Allen}.} \bibinfo{year}{2021}\natexlab{}.
\newblock \showarticletitle{Supporting learning in large classes: online formative assessment and automated feedback}. In \bibinfo{booktitle}{\emph{2021 30th {Annual} {Conference} of the {European} {Association} for {Education} in {Electrical} and {Information} {Engineering} ({EAEEIE})}}. \bibinfo{publisher}{IEEE}, \bibinfo{address}{Prague, Czech Republic}, \bibinfo{pages}{1--4}.
\newblock
\showISBNx{978-1-72819-327-4}
\urldef\tempurl%
\url{https://doi.org/10.1109/EAEEIE50507.2021.9530953}
\showDOI{\tempurl}


\bibitem[Portaz et~al\mbox{.}(2024)]%
        {portaz_exploring_2024}
\bibfield{author}{\bibinfo{person}{Miguel Portaz}, \bibinfo{person}{Alberto Corbi}, \bibinfo{person}{Alberto Casas-Ortiz}, {and} \bibinfo{person}{Olga~C. Santos}.} \bibinfo{year}{2024}\natexlab{}.
\newblock \showarticletitle{Exploring raw data transformations on inertial sensor data to model user expertise when learning psychomotor skills}.
\newblock \bibinfo{journal}{\emph{User Model User-Adap Inter}}  \bibinfo{volume}{2024} (\bibinfo{date}{April} \bibinfo{year}{2024}), \bibinfo{numpages}{43}~pages.
\newblock
\showISSN{0924-1868, 1573-1391}
\urldef\tempurl%
\url{https://doi.org/10.1007/s11257-024-09393-2}
\showDOI{\tempurl}


\bibitem[Pretolesi et~al\mbox{.}(2024)]%
        {pretolesi_ai-supported_2024}
\bibfield{author}{\bibinfo{person}{Daniele Pretolesi}, \bibinfo{person}{Olivia Zechner}, \bibinfo{person}{Daniel~Garcia Guirao}, \bibinfo{person}{Helmut Schrom-Feiertag}, {and} \bibinfo{person}{Manfred Tscheligi}.} \bibinfo{year}{2024}\natexlab{}.
\newblock \showarticletitle{{AI}-{Supported} {XR} {Training}: {Personalizing} {Medical} {First} {Responder} {Training}}. In \bibinfo{booktitle}{\emph{{AI} {Technologies} and {Virtual} {Reality}}}, \bibfield{editor}{\bibinfo{person}{Kazumi Nakamatsu}, \bibinfo{person}{Srikanta Patnaik}, {and} \bibinfo{person}{Roumen Kountchev}} (Eds.). \bibinfo{publisher}{Springer Nature Singapore}, \bibinfo{address}{Singapore}, \bibinfo{pages}{343--356}.
\newblock
\showISBNx{978-981-9990-18-4}


\bibitem[Pérez-D’Arpino et~al\mbox{.}(2023)]%
        {perez-darpino_experimental_2023}
\bibfield{author}{\bibinfo{person}{Claudia Pérez-D’Arpino}, \bibinfo{person}{Rebecca~P. Khurshid}, {and} \bibinfo{person}{Julie~A. Shah}.} \bibinfo{year}{2023}\natexlab{}.
\newblock \showarticletitle{Experimental {Assessment} of {Human}-{Robot} {Teaming} for {Multi}-{Step} {Remote} {Manipulation} with {Expert} {Operators}}.
\newblock \bibinfo{journal}{\emph{J. Hum.-Robot Interact.}}  \bibinfo{volume}{2023} (\bibinfo{date}{Oct.} \bibinfo{year}{2023}), \bibinfo{pages}{3618258}.
\newblock
\showISSN{2573-9522}
\urldef\tempurl%
\url{https://doi.org/10.1145/3618258}
\showDOI{\tempurl}


\bibitem[Pérez-Ramírez et~al\mbox{.}(2016)]%
        {perez-ramirez_intelligent_2016}
\bibfield{author}{\bibinfo{person}{Miguel Pérez-Ramírez}, \bibinfo{person}{Norma~J. Ontiveros-Hernández}, \bibinfo{person}{Carlos~A. Ochoa-Ortíz}, \bibinfo{person}{José~A. Hernández-Aguilar}, {and} \bibinfo{person}{Benjamín~E. Zayas-Pérez}.} \bibinfo{year}{2016}\natexlab{}.
\newblock \showarticletitle{Intelligent {Tutoring} {Systems} based on {Virtual} {Reality} for the {Electrical} {Domain}}.
\newblock \bibinfo{journal}{\emph{RCS}} \bibinfo{volume}{122}, \bibinfo{number}{1} (\bibinfo{date}{Dec.} \bibinfo{year}{2016}), \bibinfo{pages}{163--174}.
\newblock
\showISSN{1870-4069}
\urldef\tempurl%
\url{https://doi.org/10.13053/rcs-122-1-13}
\showDOI{\tempurl}


\bibitem[Rauter et~al\mbox{.}(2019)]%
        {rauter_when_2019}
\bibfield{author}{\bibinfo{person}{Georg Rauter}, \bibinfo{person}{Nicolas Gerig}, \bibinfo{person}{Roland Sigrist}, \bibinfo{person}{Robert Riener}, {and} \bibinfo{person}{Peter Wolf}.} \bibinfo{year}{2019}\natexlab{}.
\newblock \showarticletitle{When a robot teaches humans: {Automated} feedback selection accelerates motor learning}.
\newblock \bibinfo{journal}{\emph{Sci. Robot.}} \bibinfo{volume}{4}, \bibinfo{number}{27} (\bibinfo{date}{Feb.} \bibinfo{year}{2019}), \bibinfo{pages}{eaav1560}.
\newblock
\showISSN{2470-9476}
\urldef\tempurl%
\url{https://doi.org/10.1126/scirobotics.aav1560}
\showDOI{\tempurl}


\bibitem[Rosen et~al\mbox{.}(2011)]%
        {rosen_objective_2011}
\bibfield{author}{\bibinfo{person}{Jacob Rosen}, \bibinfo{person}{Mika Sinanan}, {and} \bibinfo{person}{Blake Hannaford}.} \bibinfo{year}{2011}\natexlab{}.
\newblock \showarticletitle{Objective {Assessment} of {Surgical} {Skills}}.
\newblock In \bibinfo{booktitle}{\emph{Surgical {Robotics}: {Systems} {Applications} and {Visions}}}, \bibfield{editor}{\bibinfo{person}{Jacob Rosen}, \bibinfo{person}{Blake Hannaford}, {and} \bibinfo{person}{Richard~M. Satava}} (Eds.). \bibinfo{publisher}{Springer US}, \bibinfo{address}{Boston, MA}, \bibinfo{pages}{619--649}.
\newblock
\showISBNx{978-1-4419-1126-1}
\urldef\tempurl%
\url{https://doi.org/10.1007/978-1-4419-1126-1_25}
\showDOI{\tempurl}


\bibitem[Santos(2016)]%
        {santos_training_2016}
\bibfield{author}{\bibinfo{person}{Olga~C. Santos}.} \bibinfo{year}{2016}\natexlab{}.
\newblock \showarticletitle{Training the {Body}: {The} {Potential} of {AIED} to {Support} {Personalized} {Motor} {Skills} {Learning}}.
\newblock \bibinfo{journal}{\emph{Int J Artif Intell Educ}} \bibinfo{volume}{26}, \bibinfo{number}{2} (\bibinfo{date}{June} \bibinfo{year}{2016}), \bibinfo{pages}{730--755}.
\newblock
\showISSN{1560-4306}
\urldef\tempurl%
\url{https://doi.org/10.1007/s40593-016-0103-2}
\showDOI{\tempurl}


\bibitem[Schwab and Zahidi(2020)]%
        {schwab_future_2020}
\bibfield{author}{\bibinfo{person}{Klaus Schwab} {and} \bibinfo{person}{Saadia Zahidi}.} \bibinfo{year}{2020}\natexlab{}.
\newblock \bibinfo{booktitle}{\emph{The {Future} of {Jobs} {Report} 2020}}.
\newblock \bibinfo{type}{{T}echnical {R}eport}. \bibinfo{institution}{World Economic Forum}.
\newblock
\urldef\tempurl%
\url{https://www3.weforum.org/docs/WEF_Future_of_Jobs_2020.pdf}
\showURL{%
\tempurl}


\bibitem[Sheehan et~al\mbox{.}(2019)]%
        {sheehan_formative_2019}
\bibfield{author}{\bibinfo{person}{Florence~H Sheehan}, \bibinfo{person}{Shannon McConnaughey}, \bibinfo{person}{Rosario Freeman}, {and} \bibinfo{person}{R~Eugene Zierler}.} \bibinfo{year}{2019}\natexlab{}.
\newblock \showarticletitle{Formative {Assessment} of {Performance} in {Diagnostic} {Ultrasound} {Using} {Simulation} and {Quantitative} and {Objective} {Metrics}}.
\newblock \bibinfo{journal}{\emph{Military Medicine}} \bibinfo{volume}{184}, \bibinfo{number}{Supplement\_1} (\bibinfo{date}{March} \bibinfo{year}{2019}), \bibinfo{pages}{386--391}.
\newblock
\showISSN{0026-4075}
\urldef\tempurl%
\url{https://doi.org/10.1093/milmed/usy388}
\showDOI{\tempurl}


\bibitem[Shute(2008)]%
        {shute_focus_2008}
\bibfield{author}{\bibinfo{person}{Valerie~J. Shute}.} \bibinfo{year}{2008}\natexlab{}.
\newblock \showarticletitle{Focus on {Formative} {Feedback}}.
\newblock \bibinfo{journal}{\emph{Review of Educational Research}} \bibinfo{volume}{78}, \bibinfo{number}{1} (\bibinfo{year}{2008}), \bibinfo{pages}{153--189}.
\newblock
\showISSN{00346543}
\urldef\tempurl%
\url{https://doi.org/10.3102/0034654307313795}
\showDOI{\tempurl}


\bibitem[Steinfeld et~al\mbox{.}(2006)]%
        {steinfeld_common_2006}
\bibfield{author}{\bibinfo{person}{Aaron Steinfeld}, \bibinfo{person}{Terrence Fong}, \bibinfo{person}{David Kaber}, \bibinfo{person}{Michael Lewis}, \bibinfo{person}{Jean Scholtz}, \bibinfo{person}{Alan Schultz}, {and} \bibinfo{person}{Michael Goodrich}.} \bibinfo{year}{2006}\natexlab{}.
\newblock \showarticletitle{Common metrics for human-robot interaction}. In \bibinfo{booktitle}{\emph{Proceedings of the 1st {ACM} {SIGCHI}/{SIGART} conference on {Human}-robot interaction}}. \bibinfo{publisher}{ACM}, \bibinfo{address}{Salt Lake City Utah USA}, \bibinfo{pages}{33--40}.
\newblock
\showISBNx{978-1-59593-294-5}
\urldef\tempurl%
\url{https://doi.org/10.1145/1121241.1121249}
\showDOI{\tempurl}


\bibitem[Wang et~al\mbox{.}(2018)]%
        {wang_toward_2018}
\bibfield{author}{\bibinfo{person}{Ziheng Wang}, \bibinfo{person}{Isabella Reed}, {and} \bibinfo{person}{Ann~Majewicz Fey}.} \bibinfo{year}{2018}\natexlab{}.
\newblock \showarticletitle{Toward {Intuitive} {Teleoperation} in {Surgery}: {Human}-{Centric} {Evaluation} of {Teleoperation} {Algorithms} for {Robotic} {Needle} {Steering}}. In \bibinfo{booktitle}{\emph{2018 {IEEE} {International} {Conference} on {Robotics} and {Automation} ({ICRA})}}. \bibinfo{publisher}{IEEE}, \bibinfo{address}{Brisbane, QLD}, \bibinfo{pages}{5799--5806}.
\newblock
\showISBNx{978-1-5386-3081-5}
\urldef\tempurl%
\url{https://doi.org/10.1109/ICRA.2018.8460729}
\showDOI{\tempurl}


\bibitem[Williamson et~al\mbox{.}(2006)]%
        {williamson_automated_2006}
\bibfield{editor}{\bibinfo{person}{David~M. Williamson}, \bibinfo{person}{Robert~J. Mislevy}, {and} \bibinfo{person}{Isaac~I. Bejar}} (Eds.). \bibinfo{year}{2006}\natexlab{}.
\newblock \bibinfo{booktitle}{\emph{Automated {Scoring} of {Complex} {Tasks} in {Computer}-{Based} {Testing}}}.
\newblock \bibinfo{publisher}{Lawrence Erlbaum}, \bibinfo{address}{Mahwah, New Jersey}.
\newblock
\showISBNx{0-8058-4634-4}


\bibitem[Yang et~al\mbox{.}(2024)]%
        {yang_adaptive_2024}
\bibfield{author}{\bibinfo{person}{Jing Yang}, \bibinfo{person}{Juan~Antonio Barragan}, \bibinfo{person}{Jason~Michael Farrow}, \bibinfo{person}{Chandru~P. Sundaram}, \bibinfo{person}{Juan~P. Wachs}, {and} \bibinfo{person}{Denny Yu}.} \bibinfo{year}{2024}\natexlab{}.
\newblock \showarticletitle{An {Adaptive} {Human}-{Robotic} {Interaction} {Architecture} for {Augmenting} {Surgery} {Performance} {Using} {Real}-{Time} {Workload} {Sensing}—{Demonstration} of a {Semi}-autonomous {Suction} {Tool}}.
\newblock \bibinfo{journal}{\emph{Hum Factors}} \bibinfo{volume}{66}, \bibinfo{number}{4} (\bibinfo{date}{April} \bibinfo{year}{2024}), \bibinfo{pages}{1081--1102}.
\newblock
\showISSN{0018-7208}
\urldef\tempurl%
\url{https://doi.org/10.1177/00187208221129940}
\showDOI{\tempurl}
\newblock
\shownote{Publisher: SAGE Publications Inc}.


\bibitem[Yuh et~al\mbox{.}(2024b)]%
        {yuh_using_2024}
\bibfield{author}{\bibinfo{person}{Madeleine~S. Yuh}, \bibinfo{person}{Ethan Rabb}, \bibinfo{person}{Adam Thorpe}, {and} \bibinfo{person}{Neera Jain}.} \bibinfo{year}{2024}\natexlab{b}.
\newblock \showarticletitle{Using {Reward} {Shaping} to {Train} {Cognitive}–based {Control} {Policies} for {Intelligent} {Tutoring} {Systems}}. In \bibinfo{booktitle}{\emph{2024 American Control Conference (ACC)}}. \bibinfo{publisher}{IEEE}, \bibinfo{address}{Toronto, ON}, \bibinfo{numpages}{8}~pages.
\newblock


\bibitem[Yuh et~al\mbox{.}(2024a)]%
        {yuh_classification_2024}
\bibfield{author}{\bibinfo{person}{Madeleine Shuhn-Tsuan Yuh}, \bibinfo{person}{Kendric~Ray Ortiz}, \bibinfo{person}{Kylie~Sue Sommer-Kohrt}, \bibinfo{person}{Meeko Oishi}, {and} \bibinfo{person}{Neera Jain}.} \bibinfo{year}{2024}\natexlab{a}.
\newblock \showarticletitle{Classification of Human Learning Stages via Kernel Distribution Embeddings}.
\newblock \bibinfo{journal}{\emph{IEEE Open Journal of Control Systems}}  \bibinfo{volume}{3} (\bibinfo{year}{2024}), \bibinfo{pages}{102--117}.
\newblock
\showISSN{2694-085X}
\urldef\tempurl%
\url{https://doi.org/10.1109/OJCSYS.2023.3348704}
\showDOI{\tempurl}


\bibitem[Zotov and Kramkowski(2023)]%
        {zotov_moving-target_2023}
\bibfield{author}{\bibinfo{person}{Vladimir Zotov} {and} \bibinfo{person}{Eric Kramkowski}.} \bibinfo{year}{2023}\natexlab{}.
\newblock \showarticletitle{Moving-{Target} {Intelligent} {Tutoring} {System} for {Marksmanship} {Training}}.
\newblock \bibinfo{journal}{\emph{Int J Artif Intell Educ}} \bibinfo{volume}{33}, \bibinfo{number}{4} (\bibinfo{date}{Dec.} \bibinfo{year}{2023}), \bibinfo{pages}{817--842}.
\newblock
\showISSN{1560-4306}
\urldef\tempurl%
\url{https://doi.org/10.1007/s40593-022-00308-z}
\showDOI{\tempurl}


\end{thebibliography}


\end{document}